\documentclass[11pt,a4paper]{article}
\usepackage[hyperref]{eacl2021}
\usepackage{times}
\usepackage{latexsym}

\usepackage{amsmath}
\usepackage{amssymb}
\usepackage{todonotes}
\usepackage{hhline}
\usepackage{multirow}
\usepackage{array}
\usepackage{lipsum}
\usepackage{stackengine}
\usepackage{xcolor}
\usepackage{enumitem}
\usepackage[labelformat=simple]{subcaption}
\renewcommand\thesubfigure{(\alph{subfigure})}

\usepackage{url}
\usetikzlibrary{arrows.meta, calc, fit, positioning}
\usetikzlibrary{calc,positioning,shapes.multipart,decorations.pathreplacing,shapes.arrows}

\definecolor{redish}{RGB}{234, 65, 91}
\definecolor{purplish}{RGB}{233, 184, 242}
\definecolor{pink}{RGB}{214, 214, 214}
\definecolor{myorange}{RGB}{247, 202, 106}
\definecolor{mygray}{RGB}{229, 229, 229}

\definecolor{greenish}{RGB}{120, 170, 95}
\definecolor{sepia}{HTML}{671800}
\definecolor{midnightblue}{HTML}{006795}
\definecolor{orangered}{HTML}{ED135A}

\newcolumntype{K}[1]{>{\centering\arraybackslash}m{#1}} 
\newcolumntype{C}[1]{>{\centering\let\newline\\\arraybackslash\hspace{0pt}}m{#1}}

\usepackage{microtype}

\aclfinalcopy 

\title{Narration Generation for Cartoon Videos}

\author{Nikos Papasarantopoulos\thanks{\hspace{0.2cm}Work done while at University of Edinburgh.} \\
  Huawei Technologies \\
  Edinburgh, UK \\
  \texttt{nikos.papasa@gmail.com} \\\And
  Shay B. Cohen \\
  ILCC, School of Informatics \\
  University of Edinburgh, UK \\
  \texttt{scohen@inf.ed.ac.uk} \\}

\date{}

\begin{document}
\maketitle
\begin{abstract}
Research on text generation from multimodal inputs has largely focused on static images, and less on video data.
In this paper, we propose a new task, \textit{narration generation}, that is complementing videos with narration texts that are to be interjected in several places.
The narrations are part of the video and contribute to the storyline unfolding in it. 
Moreover, they are context-informed, since they include information appropriate for the timeframe of video they cover, and also, do not need to include every detail shown in input scenes, as a caption would.
We collect a new dataset from the animated television series Peppa Pig. Furthermore, we formalize the task of narration generation as including two separate tasks, timing and content generation, and present a set of models on the new task.\footnote{Please contact the authors if you need access to code or data.}
\end{abstract}

\section{Introduction}
\label{sect:introduction}

\begin{figure}[t!]
    \centering
    \newcounter{row}
    \renewcommand{\thesubfigure}{(\arabic{row}\alph{subfigure})}%
    \stepcounter{row}%
    \begin{subfigure}[t]{0.23\textwidth}
        \centering
        \includegraphics[width=\textwidth]{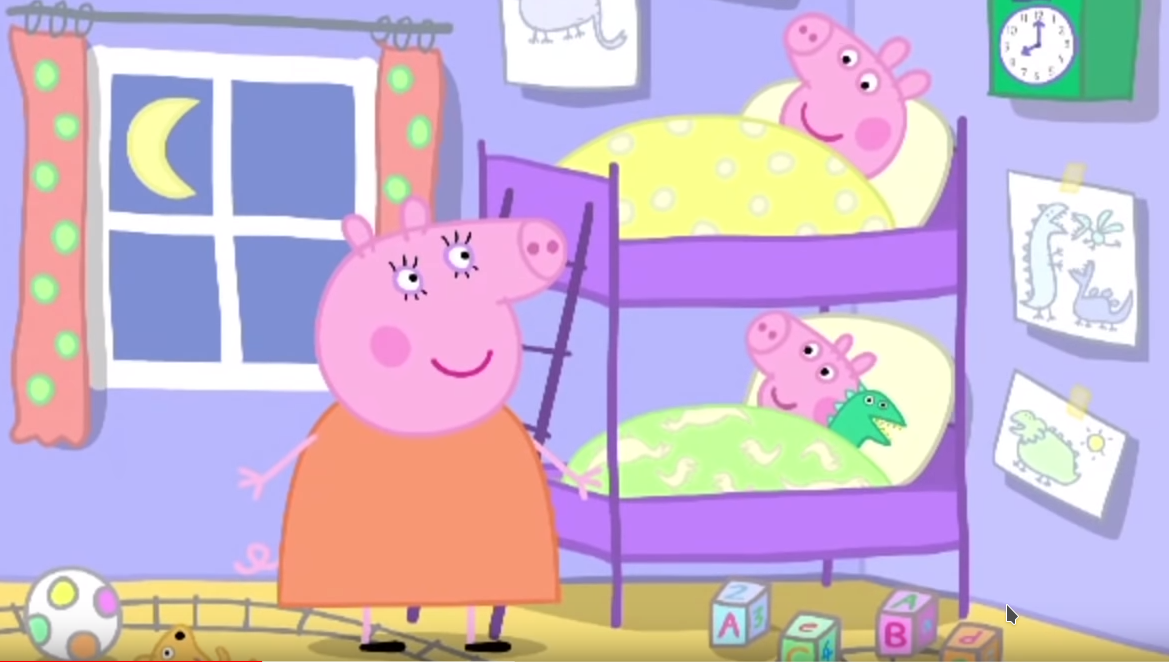}
        \caption{\centering 01:19 -- 01:20\break\textsc{Mommy Pig}: Goodnight, George.}
        \label{fig:peppa_scenes_1_a}
    \end{subfigure}\hfill%
    \begin{subfigure}[t]{0.23\textwidth}
        \centering
        \includegraphics[width=\textwidth]{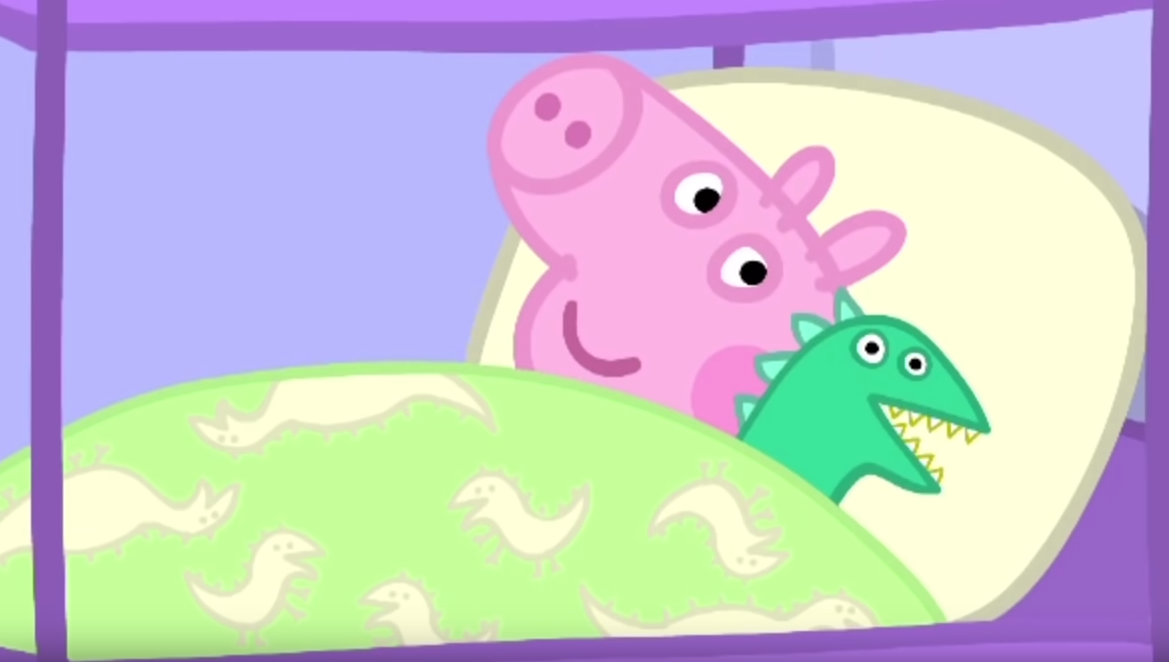}
        \caption{\centering 01:24 -- 01:27\break\textsc{Narrator}: When George goes to bed, Mr Dinosaur is tucked up with him.}
        \label{fig:peppa_scenes_1_b}
        \vspace{0.5cm}
    \end{subfigure}

    \stepcounter{row}
    \setcounter{subfigure}{0}
    \begin{subfigure}[t]{0.23\textwidth}
        \centering
        \includegraphics[width=\textwidth]{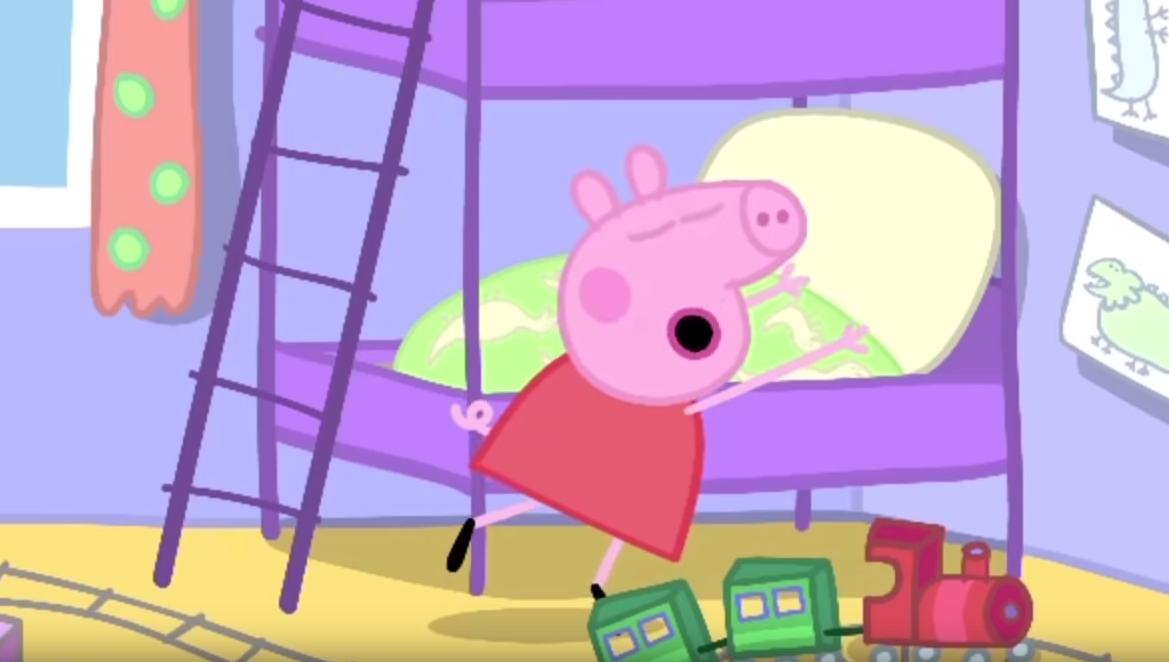}
        \caption{\centering 03:29 -- 03:31\break\textsc{Peppa Pig}: See, that's where it is.}
        \label{fig:peppa_scenes_2_a}
    \end{subfigure}\hfill%
    \begin{subfigure}[t]{0.23\textwidth}
        \centering
        \includegraphics[width=\textwidth]{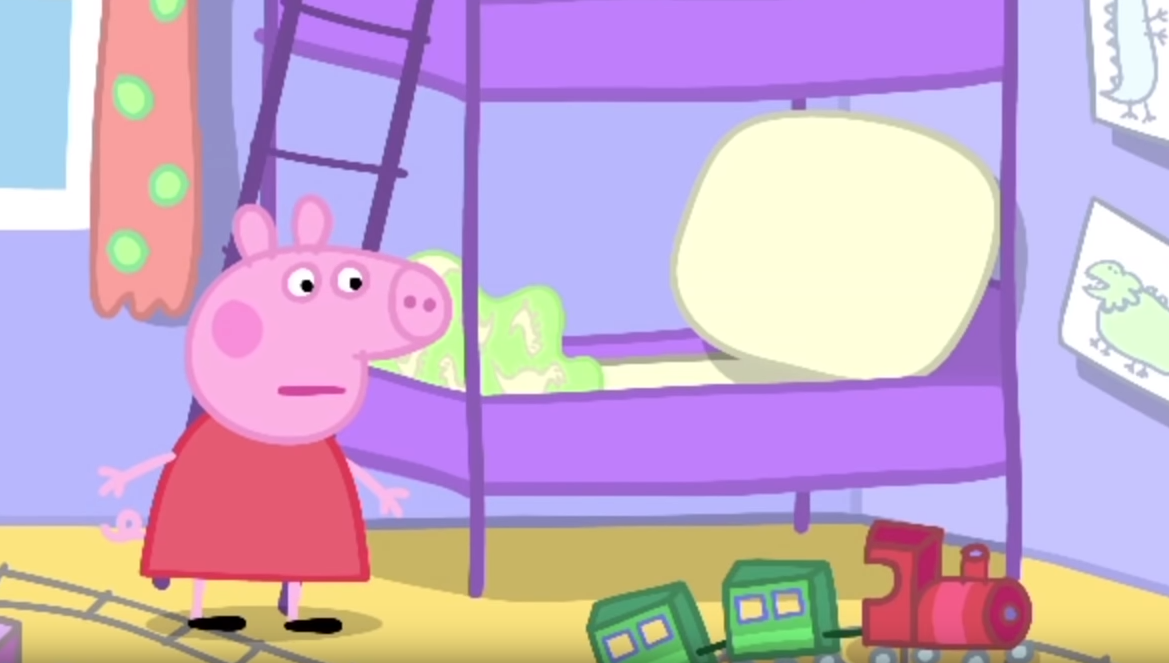}
        \caption{\centering 03:32 -- 03:34\break\textsc{Narrator}: Mr Dinosaur is not in George's bed.}
        \label{fig:peppa_scenes_2_b}
    \vspace{0.5cm}
    \end{subfigure}%

    \stepcounter{row}
    \setcounter{subfigure}{0}
    \begin{subfigure}[t]{0.23\textwidth}
        \centering
        \includegraphics[width=\textwidth]{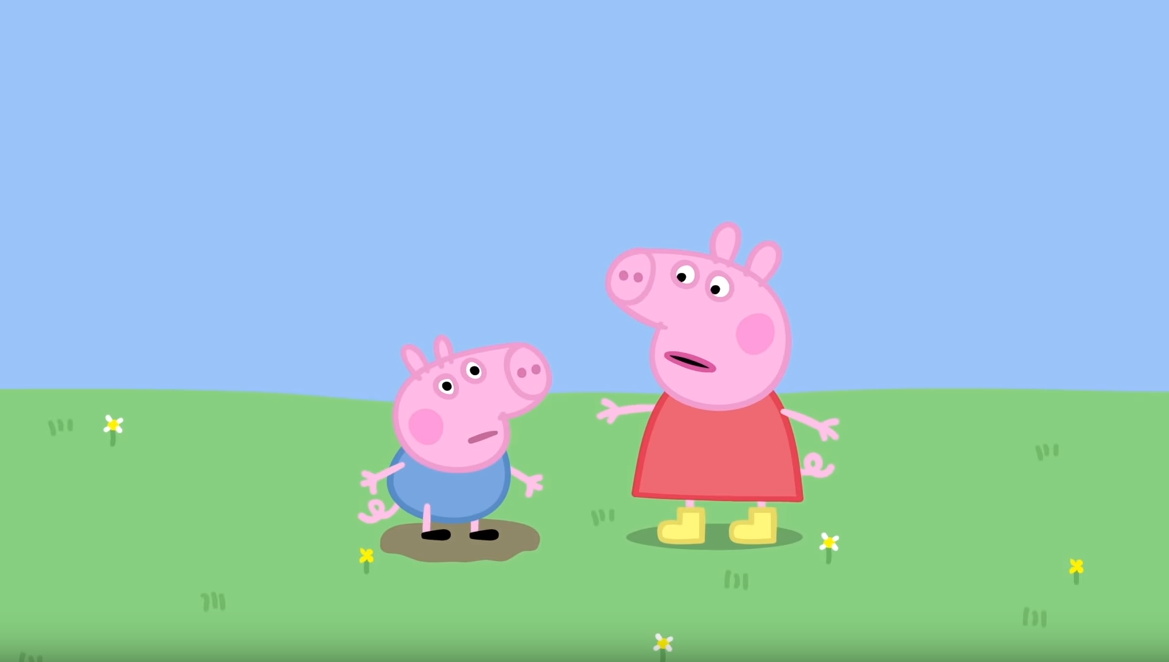}
        \caption{\centering 01:20 -- 01:26\break\textsc{Peppa Pig}: George, if you jump in muddy puddles, you must wear your boots.}
        \label{fig:peppa_scenes_3_a}
    \end{subfigure}\hfill%
    \begin{subfigure}[t]{0.23\textwidth}
        \centering
        \includegraphics[width=\textwidth]{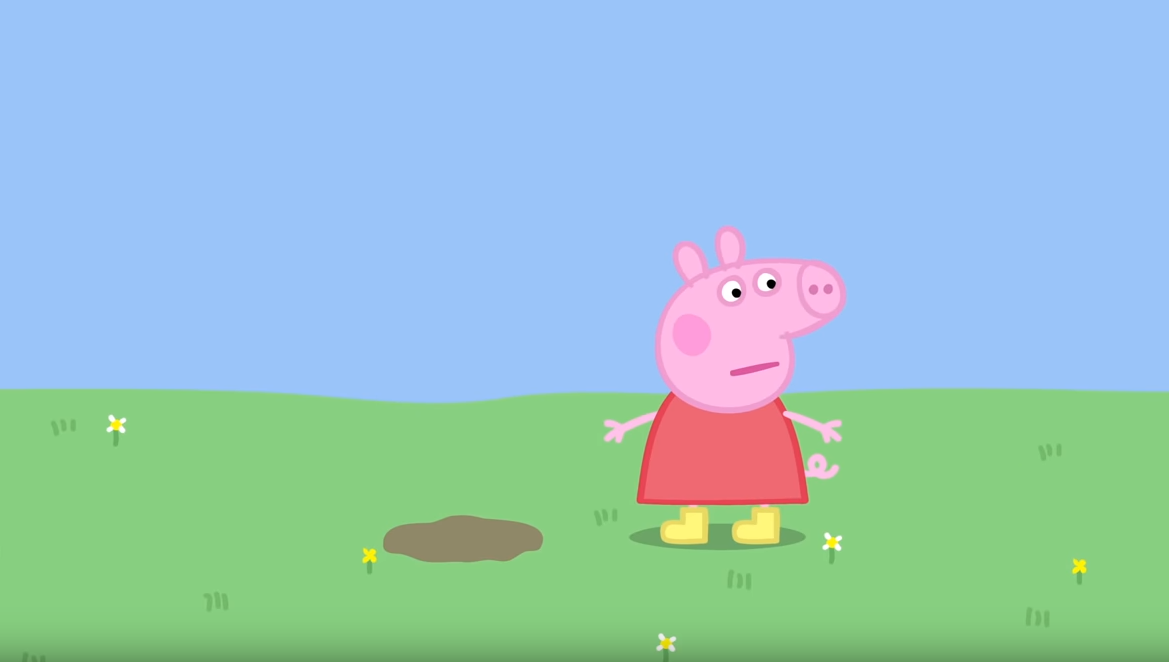}
        \caption{\centering 01:29 - 01:34\break\textsc{Narrator}: Peppa likes to look after her little brother, George.}
        \label{fig:peppa_scenes_3_b}
    \end{subfigure}%
    
\caption{An excerpt of the Peppa Pig dataset. The first two examples (rows) are from \textit{Episode 2: Mr Dinosaur is lost}, while the last from \textit{Episode 1: Muddy Puddles}. For each subtitle, a representative snapshot (image) is shown. It can be seen that the narrations may or may not be descriptive of the image (or short clip) which they accompany.
\label{fig:peppa_scenes}}\end{figure}

Text generation from visual or multimodal inputs has been an overarching goal and a point of convergence of the Computer Vision and Natural Language Processing communities \cite{Gatt2018SurveyOT}.
Research in this direction has been propelled by the proposal and study of several tasks, with examples including image caption generation (see \citealt{bernardi2016automatic} for a survey), visual question generation \cite{mostafazadeh2016generating}, caption explanation \cite{Hendricks2016GeneratingVE}, visual question answering explanation \cite{li2018vqae} and multimodal machine translation \cite{calixto2017doubly,lala-specia_LREC:2018}. 
Progress in corresponding video tasks, such as video description, video question answering \cite{zeng2017leveraging,xu2017video,MovieQA} and video overview generation \cite{gorinski2018s} has been slower, probably due to the extra challenge posed by the temporal nature of videos.
In this paper, we present \textit{narration generation}, a new text generation task from movie videos. 
We believe that the introduction of this task will challenge existing techniques for text generation from videos as it requires temporal contextual reasoning and inference on storylines.

A narration is a commentary commonly used in movies, series or books. It can be delivered by a story character, a non-personal voice or the author of a book and may communicate a story that is parallel to the plot, fill in details that are not directly perceivable, or help in guiding the viewers/readers through the plot.
\textit{Narration generation} refers to the task of automatically generating the text of such narrations. Our focus is on video data, especially videos that are part of episodic broadcasts, such as television series.

To facilitate research in the direction of automatic narration generation, we create a new narration dataset.
Following the spirit of previous work on image captioning from abstract scenes \cite{zitnick2013bringing}
and cartoon video question answering \cite{kim2016pororoqa}, we collect videos from the animated series \textit{Peppa Pig}.
We posit that abstracting away from related, but nonetheless hard problems, such as processing real-life videos and understanding complex, real-life dialogue between adults, will make for clean and isolated evaluation of the text generation techniques themselves.

Narration generation in the way we set it up, is a new task, distinct from video descriptions in certain aspects.
Importantly, narrations are not metadata; they are meant to be uttered and become part of videos. 
Compared to descriptions, narrations provide high-level, less grounded information on events taking place in videos.
In general, they do not articulate objects or actions that can be directly seen in the images and even in cases where they do, the description is tied to the context of the overall story.
For example, the scene of Figure \ref{fig:peppa_scenes_1_b}, could be accurately described by \textit{``a pig in a bed, with a toy tucked up with it"}. 
The narration, however is context-aware: it refers to the pig as ``George" and the toy as ``Mr Dinosaur".

Moreover, a narration may refer to events or objects that cannot be seen in the image.
Image captioning algorithms face striking challenges when it comes to objects absent from query images, which can be easily inferred by humans \cite{bernardi2016automatic}, such as the ``bus" in the caption of an image showing people waiting for a bus on a bench.
Narrations may include contextual mentions to absent objects, such as the reference to ``Mr Dinosaur" in Figure \ref{fig:peppa_scenes_2_b}; something that a viewer not familiar with the storyline could not have guessed.
Finally, narrations may convey information less related to their accompanying video and more to the overall plot: Figure \ref{fig:peppa_scenes_3_b}, for example, makes a remark on Peppa looking after her brother, while the image just shows Peppa near a puddle. 

Video narration generation bears resemblance to the task of automatic generation of sports broadcasts, known as sportscasts \cite{Chen2008LearningTS}.
While sportscasts can have narrative structure \cite{herman2010routledge}, they are generated on the fly, and contain information related to what has been shown in the video up to the point of their utterance. 
Conversely, the narrations of our dataset are third-person omniscient narrations: the narrator knows everything related to the storyline and may use information that is being shown to the viewers while the narration is uttered or use forward references to events that are to unfold later in the video.
In terms of content, sports commentary that does not simply describe the game, known as ``color commentary" \cite{lee2014automated} is more relevant to movie narrations.

\begin{figure}
    \centering
    \includegraphics[width=0.45\textwidth]{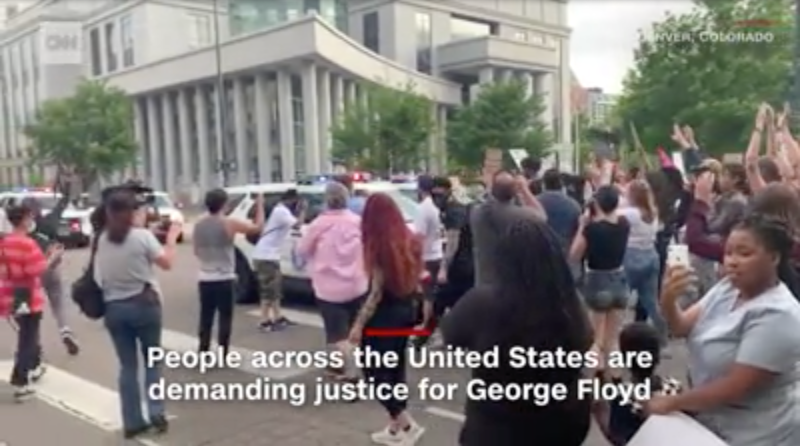}
    \caption{A snapshot from a news summary video which includes a form of narration (text shown between video excerpts to fill in gaps in the story). From the New York Times website.\footnotemark}
    \label{fig:nyt}
\end{figure}

Our work is a step towards the direction of narrative content generation and serves as a proxy problem for several applications.
Examples include not only sports, but also other types of commentary (such as director's commentary in movies) or content for news summary videos (see Figure \ref{fig:nyt}).

\footnotetext{\url{https://edition.cnn.com/us/live-news/george-floyd-protests-06-01-20/index.html}, accessed 1 June 2020, 11:00.}

The contributions of this paper are as follows:
\begin{itemize}[nosep]
    \item We introduce and formalize the task of narration generation from videos.
    \item We develop a new cartoon video dataset for the task of narration generation.
    \item We present several models for narration generation and report results on the newly introduced dataset.
\end{itemize}

\section{Dataset}
\label{sect:dataset}

\textit{Peppa Pig} is a popular British animated television series targeting preschool children. 
The episodes follow Peppa, a young female pig, in everyday social, family, and school activities. 
Main characters also include Peppa's younger brother (George), their mother (Mommy Pig), father (Daddy Pig) and various friends and relatives. 
All the friends of Peppa's family belong to different animal species than pigs. 
Although all the characters of the show are animals, most of them exhibit human traits and lead human-like lives (they speak human languages, wear clothes, live in houses, drive etc.). At the same time, they keep some animal features, such as the distinctive sound of their species (pigs, for example, snort during conversations).

The simplicity of images, dialogues and storylines of Peppa Pig, make it an ideal testbed for movie understanding experimentation.
The small scale of the vocabulary and topics alleviate sparsity challenges that can potentially arise in, relatively small but diverse, movie datasets. 
Finally, the fact that the storyline of each episode includes a narrator, makes the task of eliciting  narration data relatively easy. 

The series first aired on 2014 and as of the end of 2018, 264 episodes have been created. 
Our dataset consists of 209 episodes, for which we were able to find subtitles online. 
For each episode, we collect the video file, subtitles and metadata (title, plot summary, air date). 
Some descriptive statistics of our dataset can be found in Table \ref{table:datasetstats}, while Figure \ref{fig:peppa_scenes} lists three example scenes taken from the first two episodes of the series.
More details about narration and dialogue lengths, number and positions in episodes can be found in Appendix \ref{appendix:dataset_stats}.

\begin{table}[t]
\begin{center}
\begin{tabular}{| l | r |}
\hline
episodes & 209\\
total time & 1045 min \\
time excl. intro \& outro & 927 min\\
\hline
narrations & 1803 \\
dialogue length (avg) & 56.9 tokens \\
narration length (avg) & 10.7 tokens \\
narration vocabulary & 1771 words \\
narration unique vocabulary & 257 words \\

\hline
\end{tabular}

\end{center}
\caption{Statistics of the Peppa Pig dataset.}
\label{table:datasetstats}

\end{table}

A preprocessing step was carried out to make sure that all the videos are stripped out of unnecessary parts (each episode starts with the main character introducing herself and her family and usually ends with a song), subtitles are synchronized with the video, subtitle text is normalized, and each token of the text is aligned with a corresponding timeframe (a process called forced alignment). 
The resulting dataset is aligned at the token level, for image, audio and text modalities.
Furthermore, sentences uttered by the narrator were semi-manually annotated.
The annotation is relatively light, and does not require expert or crowdsourced annotators.
Detailed account of the preprocessing and annotation steps can be found in Appendix \ref{appendix:preprocessing}. 

\paragraph{Features} For each episode, we calculate feature vectors for the image and audio modalities, as follows. For image frames taken at the middle of each token, we use ResNet-50 (Residual Networks; \citealt{he2016deep}) and VGG-19 (from the the work of Visual Geometry Group; \citealt{simonyan2014very}).
For audio excerpts, after stripping the dialogues from the audio track,\footnote{Information on the content of the dialogues is already present in the text modality.} we calculate Mel-Frequency Cepstral Coefficients (MFCC; \citealt{mermelstein1976distance}) and VGGish features\footnote{This is the name of the pretrained model we used, available from \url{https://github.com/tensorflow/models/tree/master/research/audioset}.} \cite{hershey2017cnn,gemmeke2017audioset}, which are reported to produce state-of-the-art results in audio classification and have also been used for video description \cite{hori2018multimodal}.

\paragraph{Comparison with Other Datasets} Our dataset adds to the set of existing datasets for multimodal video understanding. 
Table \ref{table:datasetcomparison} lists features of several datasets pertaining to text generation from videos, such as video description and video question answering. 
The Peppa Pig dataset is the first dataset on narration generation. 

\begin{table*}[t]
\begin{center}
\begin{tabular}{| l | c | c | K{1cm} | K{1cm} | K{1cm} | K{1cm}|}
\hline
\textbf{dataset} & \textbf{domain} & \textbf{task} & \textbf{videos} & \textbf{avg len} (sec) & \textbf{vocab} & \textbf{len} (min) \\
\hline
MSVD \cite{chen2011collecting} & open & description & 1,970 & 10 & 13,010 & 318 \\
\hline
MPII Cooking \cite{rohrbach2012database} & cooking & description & 44 & 600 & - & 480 \\
YouCook \cite{das2013thousand} & cooking & description & 88 & - & 2,711 & 138 \\
TACoS \cite{regneri2013grounding} & cooking & description & 127 & 360 & 28,292 & 954 \\
TACoS-ML \cite{rohrbach2014coherent} & cooking & description & 185 & 360 & - & 1,626 \\
\hline
MPII-MD \cite{rohrbach2015dataset} & movie & description & 94 & 3.9 & 24,549 & 4,416 \\
M-VAD \cite{torabi2015using} & movie & description & 92 & 6.2 & 17,609 & 5,076 \\
MovieQA \cite{MovieQA} & movie & QA &  408 & 7,200 &  - & - \\
\hline
PororoQA \cite{kim2016pororoqa}& cartoon & QA & 171 & 432 & 3,729 & 1,230 \\
Peppa Pig & cartoon & narration & 209 & 300 & 1,771 & 1,045 \\
\hline
\end{tabular}
\end{center}
\caption{Multimodal datasets for tasks related to text generation from videos.}
\label{table:datasetcomparison}
\end{table*}

\paragraph{Narration as Video Summary}  
Narrations convey or hint at main events taking place during episodes. As such, they can be thought of as a form of video summary.
In order to explore whether this is true for our dataset, we  compare plot summaries and narrations using ROUGE, in order to assess the capability of an oracle narrator model as a summarizer. 
The generally low scores ($16.42$ ROUGE-1, $3.23$ ROUGE-2 F1) lead to the conclusion that narration generation is quite a distinct task from summarization, and points to the value of a dataset for this problem. Details of this evaluation and examples of episode plot summaries can be found in Appendix \ref{appendix:narration_as_summary}.

\section{Narration Generation}
\label{sect:narration_generation}

We formalize the task of narration generation as follows.
Assuming a set of $M$ videos, for which $N$ modalities are available, we regard each video as a sequence of $T$ elements $x_{ji}^k, j\in\{1,2, ..., M\}, i\in\{1, 2, ..., T\}, k\in\{1, 2, .., N\}$, where $k$ indexes the modalities and $T$ varies between videos. 
The task is to identify which elements should belong to narration and generate appropriate text for them.

The segmentation in $T$ elements can be done in three different levels: dialogue-narration (D/N), token, and time.
The first type splits the video in points where the dialogue ends and the narration begins and vice versa.\footnote{We refer to the non-narration parts of the video as dialogues, even if they do not necessarily contain dialogues.}
The second type splits the video in every token of the dialogue, and the third is a time scale, meaning that the video can be split in any timestamp. 
Our proposed models use only the first two types of segmentation.

We divide the task of narration generation in two separate tasks, \textit{timing} and \textit{content generation}, each solving a particular challenge related to it. 

\paragraph{Narration Timing} refers to the task of figuring \textit{when} to place narrations in a video.
We model timing as a tagging task, where each time step is tagged with a label indicating whether narration follows in the sequence.

Depending on the type of input data, a narration may interrupt the flow of speech in the video (as is the case with Peppa Pig: while the narrator speaks, characters do not engage in dialogue) or it may be superimposed to the rest of the speech (in sportscasts, narrations overlap with other video sounds).
In the former case, finding the timing can be regarded as an easier task, since a pause in dialogue can pinpoint the beginning of a narration. 
We model and experiment with the more general of the two, this is why we use incremental models that make predictions at each time step using information only from the previous and current time steps.
This is a common setup not only in language modeling, but also in real-time \cite{cho2016can} or multimodal reasoning applications \cite{frermann2018whodunnit}. 
The constraint of incrementality, which can be satisfied using a unidirectional Long-Short Term Memory model (LSTM; \citealt{hochreiter1997long}), is important, since look-ahead models allow information flow from future nodes, hinting the existence of narration.

\begin{figure*}[t!]
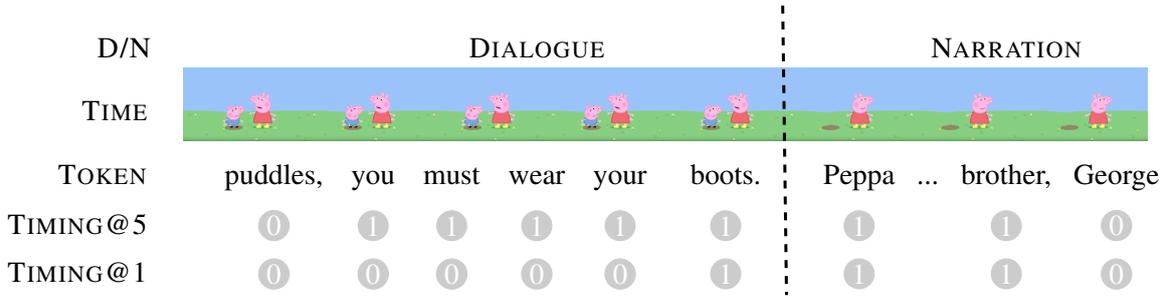

    \centering

\begin{tikzpicture}[
    ct/.style={circle, draw=white, fill=black!50,fill=black!20, text=white, inner sep=1pt, minimum width=1mm},
    noname/.style={text height=1.5ex, text depth=.25ex, text centered, minimum height=3em},
    mylabel/.style={font=\scriptsize\sffamily},
    >=LaTeX]
  \node [noname] (s1) {puddles,};
  \node [noname, right=1mm of s1] (s2) {you};
  \node [noname, left=7.5mm of s1] (s10) {\textsc{Token}};
  \node [noname, right=1mm of s2] (s3) {must};
  \node [noname, right=1mm of s3] (s4) {wear};
  \node [noname, right=1mm of s4] (s5) {your};
  \node [noname, right=3mm of s5] (s6) {boots.};
  \node [noname, right=5.5mm of s6] (s7) {Peppa};
  \node [noname, right=0.2mm of s7] (sc) {...};
  \node [noname, right=0.2mm of sc] (s8) {brother,};
  \node [noname, right=0.2mm of s8] (s9) {George};
  
  \node [noname, above left=-6mm and 1.5mm of s1] (si1) {};
  \node [noname, above left=-9mm and 3.2mm of si1] (si10) {\textsc{Time}};
  \node [noname, right=-4mm of si1] (si2) {\includegraphics[width=17mm]{peppa-george-1.png}};
  \node [noname, right=-4mm of si2] (si3) {\includegraphics[width=17mm]{peppa-george-1.png}};
  \node [noname, right=-4mm of si3] (si4) {\includegraphics[width=17mm]{peppa-george-1.png}};
  \node [noname, right=-4mm of si4] (si5) {\includegraphics[width=17mm]{peppa-george-1.png}};
  \node [noname, right=-4mm of si5] (si6) {\includegraphics[width=17mm]{peppa-george-1.png}};
  \node [noname, right=-4mm of si6] (si7) {\includegraphics[width=17mm]{peppa-george-2.png}};
  \node [noname, right=-4mm of si7] (si8) {\includegraphics[width=17mm]{peppa-george-2.png}};
  \node [noname, right=-4mm of si8] (si9) {\includegraphics[width=17mm]{peppa-george-2.png}};

  \node [ct, below=5mm of s1] (s11) {0};
  \node [noname, left=13mm of s11] (s101) {\textsc{Timing@1}};
  \node [ct, below=5mm of s2] (s21) {0};
  \node [ct, below=5mm of s3] (s31) {0};
  \node [ct, below=5mm of s4] (s41) {0};
  \node [ct, below=5mm of s5] (s51) {0};
  \node [ct, below=5mm of s6] (s61) {1};
  \node [ct, below=5mm of s7] (s71) {1};
  \node [ct, below=5mm of s8] (s81) {1};
  \node [ct, below=5mm of s9] (s91) {0};

  \node [ct, above=2mm of s11] (s12) {0};
  \node [noname, left=13mm of s12] (s101) {\textsc{Timing@5}};
  \node [ct, above=2mm of s21] (s22) {1};
  \node [ct, above=2mm of s31] (s32) {1};
  \node [ct, above=2mm of s41] (s42) {1};
  \node [ct, above=2mm of s51] (s52) {1};
  \node [ct, above=2mm of s61] (s62) {1};
  \node [ct, above=2mm of s71] (s72) {1};
  \node [ct, above=2mm of s81] (s82) {1};
  \node [ct, above=2mm of s91] (s92) {0};
  
  \node [noname, above=5mm of s4] (d) {\textsc{Dialogue}};
  \node [noname, left=39mm of d] (s10) {\textsc{D/N}};
  \node [noname, above=5mm of s8] (n) {\textsc{Narration}};

  \draw [dashed, thick, line width=0.4mm] ($(d.north west)+(4.27,-0.)$) -- ($(s7.south west)+(-0.35,-1.)$);




\end{tikzpicture}

\caption{An excerpt of an episode of the Peppa Pig dataset, where all the levels of segmentation and tagging schemes for narration timing are shown. \textit{Timing@$n$} refers to the annotation scheme, where binary labels indicate whether narration is present in a $n$-token window right after each token.
\label{fig:segmentation}}\end{figure*}

Specifically, for this task, we use the token-level segmentation of the dataset and feed an incremental sequential model with multimodal representations (tokens along with corresponding image and audio features).
Each token is annotated with a binary label, indicating whether there is at least one narration token in a window of $n$ tokens right after it.
The obvious choice is $n=1$, where each tag indicates whether the immediate next token belongs to narration.
Additionally, inspired by work on speech dialogue turn-taking \cite{skantze2017towards}, where $n>1$ is used, we create annotations with $n=5$. We refer to the two annotation schemes as \textit{Timing@1} and \textit{Timing@5}. Figure \ref{fig:segmentation} contains an annotation example from the Peppa Pig dataset.

Predicting the presence of narration in upcoming time steps is in fact a proxy of the timing problem, hence offering an upper bound on it.
Using the actual narration text (and consequently the correct tokenization, which provides correct offsets when extracting images and audio excerpts to feed to the model), makes the task easier than it really is.

\paragraph{Content Generation} refers to the task of figuring \textit{what} to include in the narrations.
In order to deal with content generation, we assume that timing is already solved and videos are correctly segmented into chunks of dialogue and narration.

We hypothesize that a human assigned with the task of coming up with a good narration for a specific part of a video would need to have access to information from several sources.
The content of the dialogue preceding the narration is of utmost importance.  
Equally important are the actions or events taking place in the part of the video that is to be narrated.
A narration generator model should have access to the same information, hence we propose a model with the following features:
\begin{itemize}[nosep]
    \item it takes into account the output of a multimodal dialogue encoder, which encodes dialogue data. 
    We instantiate a token-level LSTM encoder, which combines information from text, image and audio. 
    \item it takes into account the output of a video encoder, which encodes the part of the video to be narrated. Since the corresponding text is the desired output of our decoder, this is a video and audio decoder. Note that this encoder in principle does not have access to the narration tokenization, so it should not use the token segmentation, but a time segmentation (e.g. segment every 5ms). 
\end{itemize}

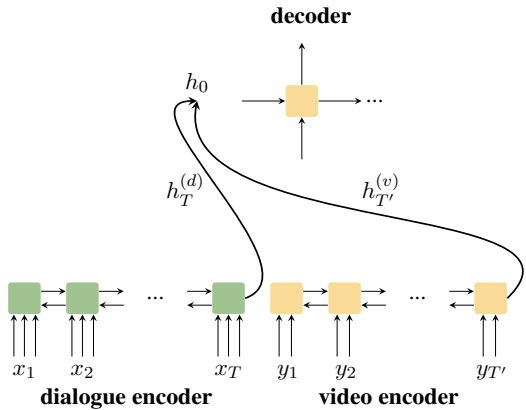
\begin{figure}[th!]
\resizebox{0.52\textwidth}{0.35\textwidth}{
\begin{tikzpicture}[
    prod/.style={circle, draw=black, inner sep=0pt, text=black,},
    ct/.style={circle, thick, draw=black, text=black, inner sep=5pt, minimum width=10mm},
    ctt/.style={circle, thick, draw=black, text=black, inner sep=5pt, minimum width=10mm},
    ft/.style={circle, thick, draw, minimum width=4mm, inner sep=1pt},
    r/.style={circle, thick, draw, minimum width=4mm, inner sep=1pt},
    rt/.style={circle, thick, draw=black, text=black, minimum width=10mm, inner sep=1pt},
    no/.style={circle, thick, draw=black, text=black, minimum width=3mm, inner sep=1pt},
    gt/.style={rectangle, thick, draw=none, text=black, minimum width=10mm, minimum height=10mm, inner sep=1pt, rounded corners=.1cm,fill=pink},
    gts/.style={rectangle, fill=myorange!70, draw=none, text=black, minimum width=15, minimum height=15, inner sep=1pt, rounded corners=0.5mm}, 
    gtsd/.style={rectangle, fill=greenish!70, draw=none, text=black, minimum width=15, minimum height=15, inner sep=1pt, rounded corners=0.5mm}, 
    gtss/.style={rectangle, thick, draw=none, text=black, minimum width=3mm, minimum height=3mm, inner sep=1pt, rounded corners=.1cm, fill=myorange},
    gtw/.style={rectangle, thick, draw=white, text=black, minimum width=10mm, minimum height=10mm, inner sep=1pt},
    filter/.style={circle, draw=black, text=black, minimum width=5mm, inner sep=1pt, path picture={\draw[thick, rounded corners] (path picture bounding box.center)--++(65:2mm)--++(0:1mm);
    \draw[thick, rounded corners] (path picture bounding box.center)--++(245:2mm)--++(180:1mm);}},
    mylabel/.style={font=\scriptsize\sffamily},
    >=LaTeX
    ]
    
\node[gtw,label={[mylabel]}] (s1) {};
\node[gtw,right=of s1, label={[mylabel]}] (s1) {};

\node[gtsd, below left=25mm and 70.00mm of s1, label={[mylabel]}] (sl1) {};
\node[gtsd, right=4mm of sl1, label={[mylabel]}] (sl2) {};
\node[gtw, right=4mm of sl2,label={[mylabel]}] (sl3) {...};
\node[gtsd, right=4mm of sl3, label={[mylabel]}] (sl4) {};

\node[gts, right=4mm of sl4, label={[mylabel]}] (dl1) {};
\node[gts, right=4mm of dl1, label={[mylabel]}] (dl2) {};
\node[gtw, right=4mm of dl2, label={[mylabel]}] (dl3) {...};
\node[gts, right=4mm of dl3, label={[mylabel]}] (dl4) {};

\draw[draw=black, fill=black, text=black,->, transform canvas={yshift=0.3em}, >=stealth] (dl1)--(dl2) node[midway, left] {};
\draw[draw=black, fill=black, text=black,<-, transform canvas={yshift=-0.3em}, >=stealth] (dl1)--(dl2) node[midway, left] {};
\draw[draw=black, fill=black, text=black,->, transform canvas={yshift=0.3em}, >=stealth] (dl2)--(dl3) node[midway, left] {};
\draw[draw=black, fill=black, text=black,<-, transform canvas={yshift=-0.3em}, >=stealth] (dl2)--(dl3) node[midway, left] {};
\draw[draw=black, fill=black, text=black,->, transform canvas={yshift=0.3em}, >=stealth] (dl3)--(dl4) node[midway, left] {};
\draw[draw=black, fill=black, text=black,<-, transform canvas={yshift=-0.3em}, >=stealth] (dl3)--(dl4) node[midway, left] {};

\node[draw, dashed, inner sep=8pt, color=white,
      fit={($(dl1.north west)+(-0.,0.4)$)  ($(dl4.south east)+(0.,-1.5)$)},
label={[label distance=-0.7cm]270:\textbf{video encoder}}] (fit) {};
\node[draw, dashed, inner sep=8pt, color=white,
      fit={($(sl1.north west)+(-0.,0.4)$)  ($(sl4.south east)+(0.,-1.5)$)},
label={[label distance=-0.7cm]270:\textbf{dialogue encoder}}] (fit) {};


%
\draw[draw=black, fill=black, text=black,->, transform canvas={yshift=0.3em}, >=stealth] (sl1)--(sl2) node[midway, left] {};
\draw[draw=black, fill=black, text=black,<-, transform canvas={yshift=-0.3em}, >=stealth] (sl1)--(sl2) node[midway, left] {};
\draw[draw=black, fill=black, text=black,->, transform canvas={yshift=0.3em}, >=stealth] (sl2)--(sl3) node[midway, left] {};
\draw[draw=black, fill=black, text=black,<-, transform canvas={yshift=-0.3em}, >=stealth] (sl2)--(sl3) node[midway, left] {};
\draw[draw=black, fill=black, text=black,->, transform canvas={yshift=0.3em}, >=stealth] (sl3)--(sl4) node[midway, left] {};
\draw[draw=black, fill=black, text=black,<-, transform canvas={yshift=-0.3em}, >=stealth] (sl3)--(sl4) node[midway, left] {};

\draw[draw=black, fill=black, text=black,<-, >=stealth] (sl1.south) coordinate (aux3)--++(270:7mm) node[below]{$x_1$};
\draw[draw=black, fill=black, text=black,<-, >=stealth] ([xshift=0.1 cm]sl1.south west) coordinate (aux3)--++(270:7mm) node[below]{};
\draw[draw=black, fill=black, text=black,<-, >=stealth] ([xshift=0.45 cm]sl1.south west) coordinate (aux3)--++(270:7mm) node[below]{};

\draw[draw=black, fill=black, text=black,<-, >=stealth] (sl2.south) coordinate (aux3)--++(270:7mm) node[below]{$x_2$};
\draw[draw=black, fill=black, text=black,<-, >=stealth] ([xshift=0.1 cm]sl2.south west) coordinate (aux3)--++(270:7mm) node[below]{};
\draw[draw=black, fill=black, text=black,<-, >=stealth] ([xshift=0.45 cm]sl2.south west) coordinate (aux3)--++(270:7mm) node[below]{};

\draw[draw=black, fill=black, text=black,<-, >=stealth] (sl4.south) coordinate (aux3)--++(270:7mm) node[below]{$x_T$};
\draw[draw=black, fill=black, text=black,<-, >=stealth] ([xshift=0.1 cm]sl4.south west) coordinate (aux3)--++(270:7mm) node[below]{};
\draw[draw=black, fill=black, text=black,<-, >=stealth] ([xshift=0.45 cm]sl4.south west) coordinate (aux3)--++(270:7mm) node[below]{};

\draw[draw=black, fill=black, text=black,<-, >=stealth] ([xshift=0.15 cm]dl1.south west) coordinate (aux3)--++(270:7mm) node[below]{$~~~y_1$};
\draw[draw=black, fill=black, text=black,<-, >=stealth] ([xshift=0.35 cm]dl1.south west) coordinate (aux3)--++(270:7mm) node[below]{};

\draw[draw=black, fill=black, text=black,<-, >=stealth] ([xshift=0.15 cm]dl2.south west) coordinate (aux3)--++(270:7mm) node[below]{$~~~y_2$};
\draw[draw=black, fill=black, text=black,<-, >=stealth] ([xshift=0.35 cm]dl2.south west) coordinate (aux3)--++(270:7mm) node[below]{};
\draw[draw=black, fill=black, text=black,<-, >=stealth] ([xshift=0.15 cm]dl4.south west) coordinate (aux3)--++(270:7mm) node[below]{$~~~y_{T'}$};
\draw[draw=black, fill=black, text=black,<-, >=stealth] ([xshift=0.35 cm]dl4.south west) coordinate (aux3)--++(270:7mm) node[below]{};

\node[gts, left=25mm of s1, label={[mylabel]}] (s4) {};
\node[gtw,left=4mm of s4, label={[mylabel]}] (sd) {};
\node[gtw,right=4mm of s4, label={[mylabel]}] (s5) {...};
\node[draw, dashed, inner sep=8pt, color=white,
      fit={($(sd.north west)+(0.1,0.7)$)  ($(s5.south east)+(0.2,-1.5)$)},
label={[label distance=-4.cm]270:\textbf{decoder}}] (fit) {};

\node[right=7mm of dl4, label={[mylabel]}] (point_for_arrow) {};
\node[above=1mm of point_for_arrow, label={[mylabel]}] (point_for_arrow_2) {};

%

\draw[draw=black, text=black,->, >=stealth, thick] (sl4.east) to [out=20,in=180] node[midway,left]{$h_{T}^{(d)}~$} (sd.west) node[above]{$h_0$};
\draw[draw=black, text=black,->, >=stealth, thick] (dl4.east) to [out=45,in=260] node[midway,above]{$h_{T'}^{(v)}$}  (sd.west) ;

\draw[draw=black, fill=black, text=black,<-, >=stealth] (s4.west) coordinate (aux3)--++(180:7mm) node[below]{};
\draw[draw=black, fill=black, text=black,->, >=stealth] (s4.north) coordinate (aux3)--++(90:7mm) node[below]{};
\draw[draw=black, fill=black, text=black,<-, >=stealth] (s4.south) coordinate (aux3)--++(270:7mm) node[below]{};
\draw[draw=black, fill=black, text=black,->, >=stealth] (s4.east) coordinate (aux3)--++(0:7mm) node[below]{};\end{tikzpicture}
}
\caption{Dialogue Video Narrator (DiViNa) model. 
Green nodes operate on dialogue (trimodal input; text, audio, image), while yellow ones on the part of the video to be narrated (bimodal input; image and audio only). 
The outputs of the two encoders are used to initialize the decoder, which generates the narration text.
}\label{figure:narration_generation} 
\end{figure}

A schematic overview of our proposed narrator model can be seen in Figure \ref{figure:narration_generation}. We call this model \textit{Dialogue Video Narrator (DiViNa)}.

\paragraph{Multimodal representation} Efficient fusion of representations from different modalities in one, multimodal representation is an open research problem and
the choice of fusion method reportedly has significant impact on downstream applications \cite{baltruvsaitis2019multimodal}.
Broadly, fusion techniques are categorized in \textit{early} (concatenation of representations at the feature level), \textit{late} (concatenation of the output of modality-specific modules) and \textit{hybrid fusion} methods \cite{atrey2010multimodal}.
Since we introduce the dataset and task, for simplicity, we employ early fusion in all our models.

\section{Experiments}
\label{sect:experiments}

This section describes our experiments for both narration timing and content generation.
This being a new task, we also present a set of baselines and report their effectiveness.

\paragraph{Experimental Setup} For all models described, we use a similar experimental setup. 
LSTMs with one layer of hidden size 500 are used in places where sequence models are needed. 
We use GloVe embeddings \cite{pennington2014glove} 
of size 300 for text, ResNet-50 features for image and the concatenation of VGGish and MFCC features for audio.\footnote{We decided on this particular combination after preliminary examination of all combinations of the features of the dataset.}
The input of multimodal modules is the concatenation of the representations of the different modalities, passed through a linear layer and a ReLU activation. The output size of this layer is 300.
We train using the Adam optimizer \cite{kingma2014adam}, with an initial learning rate of $0.001$.
During narration content generation training, we use teacher forcing \cite{williams1989learning} with probability $0.5$.
Generation is performed using a beam search decoder, with beam size of 3.

\subsection{Experiments on Timing}
We conduct experiments on timing, by training an incremental sequence tagging model, using the Timing@1 and Timing@5 annotations. 
We experiment with unimodal (text-only) and multimodal (text, audio and video) models.
The evaluation of the output of timing models is done using precision, recall and F1, as is appropriate for a tagging task. We train until the performance on the validation set stops increasing and report results on the test set.

The corresponding results can be seen in Table \ref{table:timing}.
It appears that our models do not have a particularly hard time identifying whether narration follows in the next time steps.
Interestingly, the multimodal variants outperform the unimodal ones.

\begin{table}[t]
\begin{center}
\begin{tabular}{| l | c | c | c |}
\hline
\textbf{model} & \textbf{pr} & \textbf{rec} & \textbf{f1} \\
\hline
T@1, text-only & 52.8 & \textbf{71.9} & 60.9 \\
T@1, multimodal &  \textbf{58.8} & 69.4 & \textbf{63.6}\\
\hline
T@5, text-only & 48.7 & \textbf{63.9} &  55.3  \\
T@5, multimodal & \textbf{55.3} & 61.0 &  \textbf{58.0} \\
\hline
\end{tabular}
\end{center}
\caption{Results for narration timing, using the Timing@1 (T@1) and Timing@5 (T@5) annotations. Multimodal variants use image, audio and text information.}
\label{table:timing}
\end{table}

\subsection{Experiments on Generation}

\begin{table*}[t]
\begin{center}
\begin{tabular}{| c | l | c | c | c | c | c | c |}
\hline
& \textbf{model} & \textbf{BLEU-1} & \textbf{BLEU-2} & \textbf{BLEU-3} &  \textbf{ROUGE-L} & \textbf{METEOR} & \textbf{CIDEr}\\
\hline
\multirow{3}{*}{\rotatebox[origin=c]{90}{\textsc{Text}}} & Retrieval-tfidf & 17.73 & 10.37 & 6.18 & 16.43 & 9.16 & 30.84 \\
& Retrieval-CCA & 13.45 & 7.45 & 4.74 & 12.77 & 6.39 & 21.20 \\
& Retrieval-BERT & 14.83 & 7.81 & 4.70 & 13.98 & 7.16 & 19.41 \\
\hline
\multirow{9}{*}{\rotatebox[origin=c]{90}{\textsc{Multimodal}}} & DiNa+att & 8.19 & 3.88 & 2.34 & 12.00 & 4.84 & 4.36 \\
& DiViNa  &  17.35 & 9.65 & 6.22 & 15.25 & 7.99 & 24.02 \\
& DiViNa+att & 15.70 & 8.59 & 5.82 &  13.71 & 6.88 & 25.42 \\
& Di$^2$ViNa & 17.67 & \textit{10.26} & \textit{6.76} &  \textit{15.23} & \textit{8.17} & \textit{23.61} \\
& Di$^2$ViNa+att & \textit{18.45} & 9.88 & 5.94 &  15.54 & 7.82 & 22.04 \\
\hhline{~-------}
& DiViNa+mmd & 23.20  & 13.71 & 8.95 & 20.51 & 11.15 & 49.29 \\
& DiViNa+att+mmd & 20.43  & 12.03 & 7.86 & 17.71 & 9.22 & 40.82 \\
& Di$^2$ViNa+mmd & \textbf{24.57} & \textbf{15.90} & \textbf{11.72} &  \textbf{21.70} & \textbf{12.06} & \textbf{67.24} \\
& Di$^2$ViNa+att+mmd & 22.93 & 14.79 & 10.36 &  19.96 & 10.74 & 52.93 \\
\hline
\end{tabular}
\end{center}
\caption{Results for narration content generation. The first section of the table refers to text-only baselines, while the rest of the table refers to multimodal models. The third section includes results for models using the multimodal decoder (mmd), thus having information about the desired length of the narration. The highest values are in boldface, while the highest scores for models that do not use the mmd are in italics.}
\label{table:content}

\end{table*} 

We experiment with several variations of our model and some simpler, text-only retrieval baselines.
Note that the models described in this section are trained and evaluated in dialogue/narration pairs; a different split than that used for narration timing.

The retrieval baselines work as follows: for each dialogue of the test split, instead of generating a narration from scratch, they select one from the training set, by identifying the dialogue from the training set that is closer to it. All presented retrieval methods use \textit{cosine similarity} as a similarity metric. Four retrieval baselines are tested:
\begin{itemize}[nosep]
    \item \textbf{Retrieval-tfidf} uses Term Frequency - Inverse Document Frequency (TF-IDF) representation for dialogues and narrations.
    \item \textbf{Retrieval-BERT} makes use of pre-trained Bidirectional Encoder Representations from Transformers (BERT; \citealt{devlin2019bert}) to represent both narrations and dialogues.\footnote{We used bert-as-a-service (\url{https://github.com/hanxiao/bert-as-service}) with the \texttt{BERT-Base, Uncased} pre-trained model.}
    \item \textbf{Retrieval-CCA} uses Canonical Correlation Analysis (CCA; \citealt{hotelling1935canonical}) to project the dialogue and narration spaces to a shared space.
    At test time, dialogue representations are projected to the shared space and the narrations whose projections are closest to them are selected.
    For this baseline, TF-IDF vectors are used for both dialogues and narrations.
    The CCA dimensionality is 300.\footnote{We also experimented with 100, 500 and 1000; all yielded lower performance.}
\end{itemize}    

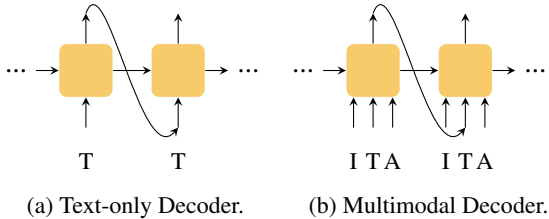
\begin{figure}
\begin{subfigure}{0.23\textwidth}
{
{%
\begin{tikzpicture}[
    gts/.style={rectangle, thick, draw=myorange, text=black, minimum width=5mm, minimum height=5mm, inner sep=1pt, rounded corners=.1cm, fill=myorange},
    rnnNode/.style={rectangle, draw=none, text=black, minimum width=7mm, minimum height=7mm, inner sep=1pt, rounded corners=.1cm, fill=myorange}, 
    gtw/.style={rectangle, thick, draw=white, text=black, minimum width=5mm, minimum height=10mm, inner sep=1pt},
    mylabel/.style={font=\scriptsize\sffamily},
    >=LaTeX
    ]

\node[rnnNode, label={[mylabel]}]  at (-200, 0) (s2) {};
\node[rnnNode, right=5mm of s2, label={[mylabel]}] (s3) {};
\node[gtw,left=3mm of s2, label={[mylabel]}] (sd) {...};
\node[gtw,right=3mm of s3, label={[mylabel]}] (sdd) {...};

\draw[draw=black, fill=black, text=black,->, >=stealth] (sd)--(s2) node[midway, below] {};;
\draw[draw=black, fill=black, text=black,->, >=stealth] (s2)--(s3) node[midway, below] {};;
\draw[draw=black, fill=black, text=black,->, >=stealth] (s3)--(sdd) node[midway, below] {};;
\draw[draw=black, fill=black, text=black,->, >=stealth] (s2.north) coordinate (aux3)--++(90:4mm) node[below]{};

\draw[draw=black, fill=black, text=black,<-, >=stealth] (s2.south) coordinate (auxOUT)--++(270:4mm) node[below, yshift=-2mm]{\small{T}};

\draw[draw=black, fill=black, text=black,->, >=stealth] (s3.north) coordinate (aux3)--++(90:4mm) node[below]{};
\draw[draw=black, fill=black, text=black,<-, >=stealth] (s3.south) coordinate (aux3)--++(270:4mm) node[below, yshift=-2mm]{\small{T}};

\draw[->, >=stealth] ([yshift=4mm]s2.north) to [out=70,in=230] ([yshift=-4mm]s3.south);
\end{tikzpicture}}}
\caption[Text-only Decoder.]{Text-only Decoder.} 
\end{subfigure}
\begin{subfigure}{0.24\textwidth}
{
\begin{tikzpicture}[
    gts/.style={rectangle, thick, draw=myorange, text=black, minimum width=5mm, minimum height=5mm, inner sep=1pt, rounded corners=.1cm, fill=myorange},
    rnnNode/.style={rectangle, draw=none, text=black, minimum width=7mm, minimum height=7mm, inner sep=1pt, rounded corners=.1cm, fill=myorange}, 
    gtw/.style={rectangle, thick, draw=white, text=black, minimum width=5mm, minimum height=10mm, inner sep=1pt},
    mylabel/.style={font=\scriptsize\sffamily},
    >=LaTeX
    ]
\node[rnnNode, label={[mylabel]}] at (-1,0) (s2) {};
\node[rnnNode, right=5mm of s2, label={[mylabel]}] (s3) {};
\node[gtw,left=3mm of s2, label={[mylabel]}] (sd) {...};
\node[gtw,right=3mm of s3, label={[mylabel]}] (sdd) {...};

\draw[draw=black, fill=black, text=black,->, >=stealth] (sd)--(s2) node[midway, below] {};;
\draw[draw=black, fill=black, text=black,->, >=stealth] (s2)--(s3) node[midway, below] {};;
\draw[draw=black, fill=black, text=black,->, >=stealth] (s3)--(sdd) node[midway, below] {};;
\draw[draw=black, fill=black, text=black,->, >=stealth] (s2.north) coordinate (aux3)--++(90:4mm) node[below]{};

\draw[draw=black, fill=black, text=black,<-, >=stealth] ([xshift=-0.1cm]s2.south east) coordinate (aux3)--++(270:4mm) node[below, yshift=-2mm]{\small{A}};
\draw[draw=black, fill=black, text=black,<-, >=stealth] (s2.south) coordinate (auxOUT)--++(270:4mm) node[below, yshift=-2mm]{\small{T}};
\draw[draw=black, fill=black, text=black,<-, >=stealth] ([xshift=0.1cm]s2.south west) coordinate (aux3)--++(270:4mm) node[below, yshift=-2mm]{\small{I}};

\draw[draw=black, fill=black, text=black,->, >=stealth] (s3.north) coordinate (aux3)--++(90:4mm) node[below]{};
\draw[draw=black, fill=black, text=black,<-, >=stealth] ([xshift=-0.1cm]s3.south east) coordinate (aux3)--++(270:4mm) node[below, yshift=-2mm]{\small{A}};
\draw[draw=black, fill=black, text=black,<-, >=stealth] (s3.south) coordinate (aux3)--++(270:4mm) node[below, yshift=-2mm]{\small{T}};
\draw[draw=black, fill=black, text=black,<-, >=stealth] ([xshift=0.1cm]s3.south west) coordinate (aux3)--++(270:4mm) node[below, yshift=-2mm]{\small{I}};

\draw[->, >=stealth] ([yshift=4mm]s2.north) to [out=70,in=230] ([yshift=-4mm]s3.south);
\end{tikzpicture}}
\caption[Multimodal Decoder.]{Multimodal Decoder.} 
\end{subfigure}
\caption{Multimodal Decoder used in \textit{mmd} variants of \textit{DiViNa} and comparison with Text-only decoder. 
At each timestep, audio (A) and image (I) representations are fed to the decoder, along with the word embedding of the corresponding token (T). In this case, the previously generated token is used (no teacher forcing).
}\label{figure:mmdecoder} 
\end{figure}

We compare the output of the baselines with that of several variants of the \textit{DiViNa} model:
\begin{itemize}[nosep]
    \item \textbf{DiViNa} is the model depicted on Figure \ref{figure:narration_generation}.
    The dialogue encoder takes as input the concatenation of text, audio and image information at each timestep $t$:
    \begin{align*}
        x_t = \mathrm{ReLU}(W [x_t^{(T)};x_t^{(A)};x_t^{(I)}]).
    \end{align*}
    The video encoder operates on the part of the video to be narrated; its input is image and audio at each timestep $t'$:
    \begin{align*}
        y_{t'} = \mathrm{ReLU}(W [y_{t'}^{(I)};y_{t'}^{(A)}]).
    \end{align*}
    The hidden states of the last elements of those encoders, $h_T^{(d)}, h_{T'}^{(v)}$ respectively, are used to initialise the decoder, after being passed through a linear layer to reduce its size:
    \begin{align*}
        h_0 = \mathrm{ReLU}(W [h_T^{(d)};h_{T'}^{(v)}]).
    \end{align*}

    \item \textbf{DiViNa+att} is the \textit{DiViNa} model equipped with an attention mechanism over the states of the dialogue encoder. For this and other attentive variants, we make use of dot product attention \cite{luong-pham-manning:2015:EMNLP}, where the attention scores between encoder ($h_i$) and decoder ($h_t^{(dec)}$) hidden states are calculated as follows:
    \begin{align*}
        \mathrm{score}(h_t^{(dec)}, h_i) = {h_t^{(dec)}}^\top h_i.
    \end{align*}

    \item \textbf{DiNa+att} is a variant of the \textit{DiViNa+att} model, with no video encoder. The narration is generated based on the preceding dialogue only; hence, the decoder is initialised with 
    \begin{align*}
        h_0 = \mathrm{ReLU}(W h_T^{(d)}).
    \end{align*}

    \item \textbf{Di$^2$ViNa, Di$^2$ViNa+att}: Motivated by the omniscience of the narrator in the dataset, these models use the dialogue that follows the video to be narrated.
    This is achieved by encoding the dialogue following the narration using a separate dialogue encoder and feeding its output $h_{T''}^{(fd)}$ to the decoder:
    \begin{align*}
        h_0 = \mathrm{ReLU}(W [h_T^{(d)};h_{T'}^{(v)};h_{T''}^{(fd)}]).
    \end{align*}

    The \textit{Di$^2$ViNa} variants are the only models that look ahead in time while generating narrations, thus being able to incorporate forward reference information.
    \textit{DiViNa} variants use only past and present information; a setup consistent with sports commentary.
    \item \textbf{DiViNa+mmd}: In order to be able to use the audio and image representations which are available for the narration part, we use a multimodal decoder (mmd).
    The basic function of such a decoder is shown in Figure \ref{figure:mmdecoder}.
    During training, the fusion of image, audio and text representations is given to the decoder at each time step. Even when not using teacher forcing (i.e. when feeding the predictions of the previous step to the next step), the input of the decoder is the concatenation of the ``correct" audiovisual representations and the word embedding of the token generated by the previous step.
    At test time, decoding proceeds for as many steps as the number of audiovisual representations available.
\end{itemize}

An obvious limitation of mmd is that it cannot generate narrations of arbitrary length. 
This is not necessarily a constraint though, since some narration setups call for narrations of predefined length, as is the case for the Peppa Pig dataset.
Since multimodal information is necessary at each timestep, quantization of the video chunk to be narrated should be done before the start of the decoding process.
The known quantization assumption is quite fair, since an estimation of the length of the narration will most likely be available in all situations (for example, it can be calculated by dividing the time between two dialogues with the mean time it takes a narrator to utter a word).
In our experiments, we exploit the known tokenization of narrations to extract multimodal representations to feed to the decoder.
As such, this model variant has access to more information than its counterparts.

In all models containing a video encoder for the narration part, we made the simplifying choice to use the correct tokenization of the narration from the dataset, despite the fact that, normally, a narrator model will not have access to that information.

Narration generation is a natural language generation (NLG) task, and as such, its evaluation is not straightforward.
Automatic evaluation of NLG systems is an open problem and widely used metrics are under heavy criticism by the community \cite{callison2006re}.
Following previous work on NLG, ranging from image captioning \cite{xu2015show} to summarization \cite{see2017get}, we report a set of word overlap metrics, complementary to each other.\footnote{We used \texttt{nlg-eval}: \texttt{\url{https://github.com/Maluuba/nlg-eval}}.}
In the case of narration generation, these metrics compare narrations from the dataset with those generated by our systems, without considering the videos.
The scores for all models can be seen in Table \ref{table:content}.





The generally low scores of the retrieval models suggest that the task of narration generation is not trivial and highlights the need for more sophisticated generative models.
Quite interestingly though, the simple TF-IDF retrieval baseline not only outperforms other text-only retrieval baselines, but also performs on par with some of the simplest generation multimodal models.

The significantly lower performance of \textit{DiNa+att} suggests that encoding the video of the part to be narrated is of high importance. Equally important is encoding the upcoming dialogue, as suggested by the generally higher scores of the Di$^2$ViNa variants.
The attentive variants of all models seem to perform lower than their non-attentive counterparts. 
A reason for that may be that they base their predictions more on the preceding dialogue (since they attend to it), while the contributions of the other encoders are downgraded.
These results highlight a distinctive quality of the task of narration generation: the significance of effective combination of information from the past (preceding dialogue), the present (video to be narrated) and the future (upcoming dialogue).

The fact that variants using the multimodal decoder (mmd) outperform all other variants does not come as a surprise, since they have access to more information during decoding and also generate narrations of the right length, by design. Examples of the generated narrations are shown in Figure \ref{fig:examples}.

\begin{figure}[th!]
    \centering
    \begin{tabular}{| p{0.45\textwidth} | }
\hline
\small \textsc{\textbf{M:}} Peppa and jumping up and down in muddy puddles. Everyone loves jumping up. \\
\small \textsc{\textbf{GT:}} Everyone in the whole world loves jumping up and down in muddy puddles. \\
\hline
\small \textsc{\textbf{M:}} Suzy sheep has come to play on the. \\
\small \textsc{\textbf{GT:}} Peppa has come to play with Suzy sheep.\\
\hline
\small \textsc{\textbf{M:}} It is bedtime for Peppa and George are very sleepy.\\
\small \textsc{\textbf{GT:}} It is nighttime. Peppa and George are going to bed.\\
\hline
\end{tabular}
\caption{Examples of output of Di$^2$ViNa+mmd, paired with respective ground truth narrations (M stands for model, GT for ground truth).
\label{fig:examples}}\end{figure}

\section{Conclusions}
\label{sect:conclusions}

In this paper, we introduce and formalize the task of narration generation from videos.
Narration generation refers to the task of accompanying videos with text snippets in several places; text that is meant to be uttered by a narrator and become part of the video.
The task of narration generation adds to the set of research tasks related to text generation from multimodal input.
The problem includes a timing (figuring out when to narrate) and a content generation (figuring out what to include in the narration) part.
Due to this dual nature of the problem, along with the fact that the content of narrations is, in general, context-aware and not particularly descriptive of the images shown in the video, we believe that this is a new and challenging task. 

To facilitate research in the direction of automatic narration generation, we create a dataset from the animated television series Peppa Pig, whose episodes include narration.
We propose neural models to tackle both problems of narration timing and content generation and report the first results on the newly introduced task and dataset.

\section*{Acknowledgments}
The authors would like to thank Esma Balk\i r, Marco Damonte, Lucia Specia and Ivan Titov for their help and comments. This research was supported by a grant from the European Union H2020 project SUMMA, under grant agreement 688139.

\bibliographystyle{acl_natbib}
\bibliography{eacl2021}

\begin{thebibliography}{46}
\expandafter\ifx\csname natexlab\endcsname\relax\def\natexlab#1{#1}\fi

\bibitem[{Atrey et~al.(2010)Atrey, Hossain, El~Saddik, and
  Kankanhalli}]{atrey2010multimodal}
Pradeep~K Atrey, M~Anwar Hossain, Abdulmotaleb El~Saddik, and Mohan~S
  Kankanhalli. 2010.
\newblock Multimodal fusion for multimedia analysis: a survey.
\newblock \emph{Multimedia systems}, 16(6):345--379.

\bibitem[{Baltru{\v{s}}aitis et~al.(2019)Baltru{\v{s}}aitis, Ahuja, and
  Morency}]{baltruvsaitis2019multimodal}
Tadas Baltru{\v{s}}aitis, Chaitanya Ahuja, and Louis-Philippe Morency. 2019.
\newblock Multimodal machine learning: A survey and taxonomy.
\newblock \emph{IEEE Transactions on Pattern Analysis and Machine
  Intelligence}, 41(2):423--443.

\bibitem[{Bernardi et~al.(2016)Bernardi, Cakici, Elliott, Erdem, Erdem,
  Ikizler-Cinbis, Keller, Muscat, and Plank}]{bernardi2016automatic}
Raffaella Bernardi, Ruket Cakici, Desmond Elliott, Aykut Erdem, Erkut Erdem,
  Nazli Ikizler-Cinbis, Frank Keller, Adrian Muscat, and Barbara Plank. 2016.
\newblock Automatic description generation from images: A survey of models,
  datasets, and evaluation measures.
\newblock \emph{JAIR}, 55:409--442.

\bibitem[{Calixto et~al.(2017)Calixto, Liu, and Campbell}]{calixto2017doubly}
Iacer Calixto, Qun Liu, and Nick Campbell. 2017.
\newblock Doubly-attentive decoder for multi-modal neural machine translation.
\newblock In \emph{Proc. of ACL}.

\bibitem[{Callison-Burch et~al.(2006)Callison-Burch, Osborne, and
  Koehn}]{callison2006re}
Chris Callison-Burch, Miles Osborne, and Philipp Koehn. 2006.
\newblock Re-evaluation the role of bleu in machine translation research.
\newblock In \emph{EACL}.

\bibitem[{Chen and Dolan(2011)}]{chen2011collecting}
David~L Chen and William~B Dolan. 2011.
\newblock Collecting highly parallel data for paraphrase evaluation.
\newblock In \emph{Proc. of ACL}.

\bibitem[{Chen and Mooney(2008)}]{Chen2008LearningTS}
David~L. Chen and Raymond~J. Mooney. 2008.
\newblock Learning to sportscast: a test of grounded language acquisition.
\newblock In \emph{ICML}.

\bibitem[{Cho and Esipova(2016)}]{cho2016can}
Kyunghyun Cho and Masha Esipova. 2016.
\newblock Can neural machine translation do simultaneous translation?
\newblock \emph{arXiv:1606.02012}.

\bibitem[{Das et~al.(2013)Das, Xu, Doell, and Corso}]{das2013thousand}
Pradipto Das, Chenliang Xu, Richard~F Doell, and Jason~J Corso. 2013.
\newblock A thousand frames in just a few words: Lingual description of videos
  through latent topics and sparse object stitching.
\newblock In \emph{Proc. of CVPR}.

\bibitem[{Devlin et~al.(2019)Devlin, Chang, Lee, and
  Toutanova}]{devlin2019bert}
Jacob Devlin, Ming-Wei Chang, Kenton Lee, and Kristina Toutanova. 2019.
\newblock {BERT}: Pre-training of deep bidirectional transformers for language
  understanding.
\newblock In \emph{Proc. of NAACL-HLT}.

\bibitem[{Frermann et~al.(2018)Frermann, Cohen, and
  Lapata}]{frermann2018whodunnit}
Lea Frermann, Shay~B Cohen, and Mirella Lapata. 2018.
\newblock Whodunnit? crime drama as a case for natural language understanding.
\newblock \emph{TACL}, 6:1--15.

\bibitem[{Gatt and Krahmer(2018)}]{Gatt2018SurveyOT}
Albert Gatt and Emiel Krahmer. 2018.
\newblock Survey of the state of the art in natural language generation: Core
  tasks, applications and evaluation.
\newblock \emph{JAIR}, 61:65--170.

\bibitem[{Gemmeke et~al.(2017)Gemmeke, Ellis, Freedman, Jansen, Lawrence,
  Moore, Plakal, and Ritter}]{gemmeke2017audioset}
Jort~F. Gemmeke, Daniel P.~W. Ellis, Dylan Freedman, Aren Jansen, Wade
  Lawrence, R.~Channing Moore, Manoj Plakal, and Marvin Ritter. 2017.
\newblock Audio set: An ontology and human-labeled dataset for audio events.
\newblock In \emph{Proc. IEEE ICASSP 2017}.

\bibitem[{Gorinski and Lapata(2018)}]{gorinski2018s}
Philip~John Gorinski and Mirella Lapata. 2018.
\newblock What’s this movie about? a joint neural network architecture for
  movie content analysis.
\newblock In \emph{Proc. of NAACL-HLT}.

\bibitem[{He et~al.(2016)He, Zhang, Ren, and Sun}]{he2016deep}
Kaiming He, Xiangyu Zhang, Shaoqing Ren, and Jian Sun. 2016.
\newblock Deep residual learning for image recognition.
\newblock In \emph{Proc. of CVPR}.

\bibitem[{Hendricks et~al.(2016)Hendricks, Akata, Rohrbach, Donahue, Schiele,
  and Darrell}]{Hendricks2016GeneratingVE}
Lisa~Anne Hendricks, Zeynep Akata, Marcus Rohrbach, Jeff Donahue, Bernt
  Schiele, and Trevor Darrell. 2016.
\newblock Generating visual explanations.
\newblock In \emph{ECCV}.

\bibitem[{Herman et~al.(2010)Herman, Manfred, and
  Marie-Laure}]{herman2010routledge}
David Herman, JAHN Manfred, and RYAN Marie-Laure. 2010.
\newblock \emph{Routledge encyclopedia of narrative theory}.
\newblock Routledge.

\bibitem[{Hershey et~al.(2017)Hershey, Chaudhuri, Ellis, Gemmeke, Jansen,
  Moore, Plakal, Platt, Saurous, Seybold, Slaney, Weiss, and
  Wilson}]{hershey2017cnn}
Shawn Hershey, Sourish Chaudhuri, Daniel~PW Ellis, Jort~F Gemmeke, Aren Jansen,
  R~Channing Moore, Manoj Plakal, Devin Platt, Rif~A Saurous, Bryan Seybold,
  Malcolm Slaney, Ron~J Weiss, and Kevin~W Wilson. 2017.
\newblock Cnn architectures for large-scale audio classification.
\newblock In \emph{Proc. of ICASSP}.

\bibitem[{Hochreiter and Schmidhuber(1997)}]{hochreiter1997long}
Sepp Hochreiter and J{\"u}rgen Schmidhuber. 1997.
\newblock Long short-term memory.
\newblock \emph{Neural computation}, 9(8):1735--1780.

\bibitem[{Hori et~al.(2018)Hori, Hori, Wichern, Wang, Lee, Cherian, and
  Marks}]{hori2018multimodal}
Chiori Hori, Takaaki Hori, Gordon Wichern, Jue Wang, Teng-yok Lee, Anoop
  Cherian, and Tim~K Marks. 2018.
\newblock Multimodal attention for fusion of audio and spatiotemporal features
  for video description.
\newblock In \emph{Proc. CVPR Workshops}.

\bibitem[{Hotelling(1935)}]{hotelling1935canonical}
Harold Hotelling. 1935.
\newblock Canonical correlation analysis (cca).
\newblock \emph{Journal of Educational Psychology}.

\bibitem[{Kim et~al.(2016)Kim, Heo, and Zhang}]{kim2016pororoqa}
KyungMin Kim, Min-Oh Heo, and Byoung-Tak Zhang. 2016.
\newblock Pororoqa: Cartoon video series dataset for story understanding.
\newblock In \emph{Proc. of NIPS Workshop on Large Scale Computer Vision
  System}.

\bibitem[{Kingma and Ba(2014)}]{kingma2014adam}
Diederik~P Kingma and Jimmy Ba. 2014.
\newblock Adam: A method for stochastic optimization.
\newblock \emph{arXiv:1412.6980}.

\bibitem[{Lala and Specia(2018)}]{lala-specia_LREC:2018}
Chiraag Lala and Lucia Specia. 2018.
\newblock Multimodal lexical translation.
\newblock In \emph{Proc. of LREC}.

\bibitem[{Lee et~al.(2014)Lee, Bulitko, and Ludvig}]{lee2014automated}
Greg Lee, Vadim Bulitko, and Elliot~A Ludvig. 2014.
\newblock Automated story selection for color commentary in sports.
\newblock \emph{T-CIAIG}, 6(2):144--155.

\bibitem[{Li et~al.(2018)Li, Tao, Joty, Cai, and Luo}]{li2018vqae}
Qing Li, Qingyi Tao, Shafiq Joty, Jianfei Cai, and Jiebo Luo. 2018.
\newblock Vqa-e: Explaining, elaborating, and enhancing your answers for visual
  questions.
\newblock In \emph{ECCV}.

\bibitem[{Lin(2004)}]{lin2004rouge}
Chin-Yew Lin. 2004.
\newblock Rouge: A package for automatic evaluation of summaries.
\newblock In \emph{Text Summarization Branches Out}.

\bibitem[{Luong et~al.(2015)Luong, Pham, and
  Manning}]{luong-pham-manning:2015:EMNLP}
Minh-Thang Luong, Hieu Pham, and Christopher~D. Manning. 2015.
\newblock Effective approaches to attention-based neural machine translation.
\newblock In \emph{Proc. of EMNLP}.

\bibitem[{Mermelstein(1976)}]{mermelstein1976distance}
Paul Mermelstein. 1976.
\newblock Distance measures for speech recognition, psychological and
  instrumental.
\newblock \emph{Pattern recognition and artificial intelligence}, 116:374--388.

\bibitem[{Mostafazadeh et~al.(2016)Mostafazadeh, Misra, Devlin, Mitchell, He,
  and Vanderwende}]{mostafazadeh2016generating}
Nasrin Mostafazadeh, Ishan Misra, Jacob Devlin, Margaret Mitchell, Xiaodong He,
  and Lucy Vanderwende. 2016.
\newblock Generating natural questions about an image.
\newblock In \emph{Proc. of ACL}.

\bibitem[{Pennington et~al.(2014)Pennington, Socher, and
  Manning}]{pennington2014glove}
Jeffrey Pennington, Richard Socher, and Christopher Manning. 2014.
\newblock Glove: Global vectors for word representation.
\newblock In \emph{Proc. of EMNLP}.

\bibitem[{Regneri et~al.(2013)Regneri, Rohrbach, Wetzel, Thater, Schiele, and
  Pinkal}]{regneri2013grounding}
Michaela Regneri, Marcus Rohrbach, Dominikus Wetzel, Stefan Thater, Bernt
  Schiele, and Manfred Pinkal. 2013.
\newblock Grounding action descriptions in videos.
\newblock \emph{TACL}, 1:25--36.

\bibitem[{Rohrbach et~al.(2014)Rohrbach, Rohrbach, Qiu, Friedrich, Pinkal, and
  Schiele}]{rohrbach2014coherent}
Anna Rohrbach, Marcus Rohrbach, Wei Qiu, Annemarie Friedrich, Manfred Pinkal,
  and Bernt Schiele. 2014.
\newblock Coherent multi-sentence video description with variable level of
  detail.
\newblock In \emph{Proc. of GCPR}.

\bibitem[{Rohrbach et~al.(2015)Rohrbach, Rohrbach, Tandon, and
  Schiele}]{rohrbach2015dataset}
Anna Rohrbach, Marcus Rohrbach, Niket Tandon, and Bernt Schiele. 2015.
\newblock A dataset for movie description.
\newblock In \emph{Proc. of CVPR}.

\bibitem[{Rohrbach et~al.(2012)Rohrbach, Amin, Andriluka, and
  Schiele}]{rohrbach2012database}
Marcus Rohrbach, Sikandar Amin, Mykhaylo Andriluka, and Bernt Schiele. 2012.
\newblock A database for fine grained activity detection of cooking activities.
\newblock In \emph{Proc. of CVPR}.

\bibitem[{See et~al.(2017)See, Liu, and Manning}]{see2017get}
Abigail See, Peter~J Liu, and Christopher~D Manning. 2017.
\newblock Get to the point: Summarization with pointer-generator networks.
\newblock In \emph{Proc. of ACL}.

\bibitem[{Simonyan and Zisserman(2014)}]{simonyan2014very}
Karen Simonyan and Andrew Zisserman. 2014.
\newblock Very deep convolutional networks for large-scale image recognition.
\newblock \emph{arXiv:1409.1556}.

\bibitem[{Skantze(2017)}]{skantze2017towards}
Gabriel Skantze. 2017.
\newblock Towards a general, continuous model of turn-taking in spoken dialogue
  using lstm recurrent neural networks.
\newblock In \emph{Proc. of the 18th Annual SIGdial Meeting on Discourse and
  Dialogue}.

\bibitem[{Tapaswi et~al.(2016)Tapaswi, Zhu, Stiefelhagen, Torralba, Urtasun,
  and Fidler}]{MovieQA}
Makarand Tapaswi, Yukun Zhu, Rainer Stiefelhagen, Antonio Torralba, Raquel
  Urtasun, and Sanja Fidler. 2016.
\newblock {MovieQA: Understanding Stories in Movies through
  Question-Answering}.
\newblock In \emph{Proc. of CVPR}.

\bibitem[{Torabi et~al.(2015)Torabi, Pal, Larochelle, and
  Courville}]{torabi2015using}
Atousa Torabi, Christopher Pal, Hugo Larochelle, and Aaron Courville. 2015.
\newblock Using descriptive video services to create a large data source for
  video annotation research.
\newblock \emph{arXiv:1503.01070}.

\bibitem[{Williams and Zipser(1989)}]{williams1989learning}
Ronald~J Williams and David Zipser. 1989.
\newblock A learning algorithm for continually running fully recurrent neural
  networks.
\newblock \emph{Neural computation}, 1(2):270--280.

\bibitem[{Xu et~al.(2017)Xu, Zhao, Xiao, Wu, Zhang, He, and
  Zhuang}]{xu2017video}
Dejing Xu, Zhou Zhao, Jun Xiao, Fei Wu, Hanwang Zhang, Xiangnan He, and Yueting
  Zhuang. 2017.
\newblock Video question answering via gradually refined attention over
  appearance and motion.
\newblock In \emph{Proc. of the 2017 ACM on Multimedia Conference}.

\bibitem[{Xu et~al.(2015)Xu, Ba, Kiros, Cho, Courville, Salakhudinov, Zemel,
  and Bengio}]{xu2015show}
Kelvin Xu, Jimmy Ba, Ryan Kiros, Kyunghyun Cho, Aaron Courville, Ruslan
  Salakhudinov, Rich Zemel, and Yoshua Bengio. 2015.
\newblock Show, attend and tell: Neural image caption generation with visual
  attention.
\newblock In \emph{ICML}.

\bibitem[{Yuan and Liberman(2008)}]{yuan2008speaker}
Jiahong Yuan and Mark Liberman. 2008.
\newblock Speaker identification on the scotus corpus.
\newblock \emph{Journal of the Acoustical Society of America}, 123(5):3878.

\bibitem[{Zeng et~al.(2017)Zeng, Chen, Chuang, Liao, Niebles, and
  Sun}]{zeng2017leveraging}
Kuo-Hao Zeng, Tseng-Hung Chen, Ching-Yao Chuang, Yuan-Hong Liao, Juan~Carlos
  Niebles, and Min Sun. 2017.
\newblock Leveraging video descriptions to learn video question answering.
\newblock In \emph{AAAI}.

\bibitem[{Zitnick and Parikh(2013)}]{zitnick2013bringing}
C~Lawrence Zitnick and Devi Parikh. 2013.
\newblock Bringing semantics into focus using visual abstraction.
\newblock In \emph{Proc. of CVPR}.

\end{thebibliography}

\appendix

\newpage

\section{Dataset Characteristics}\label{appendix:dataset_stats}

\begin{figure}[h]
    \centering
    \begin{subfigure}[t]{0.23\textwidth}
    \includegraphics[width=\textwidth]{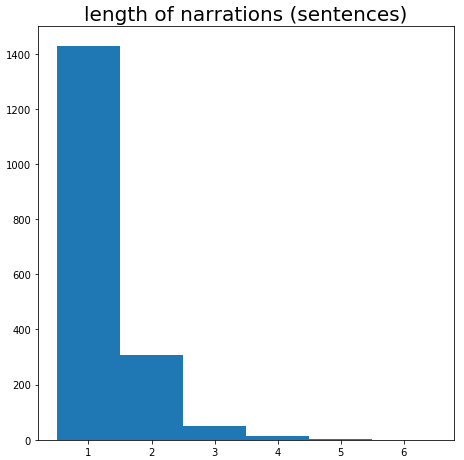}
    \end{subfigure}
    \begin{subfigure}[t]{0.23\textwidth}
    \includegraphics[width=\textwidth]{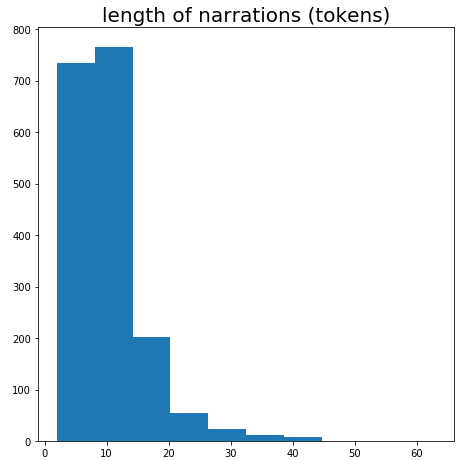}
    \end{subfigure}
    \begin{subfigure}[t]{0.23\textwidth}
    \includegraphics[width=\textwidth]{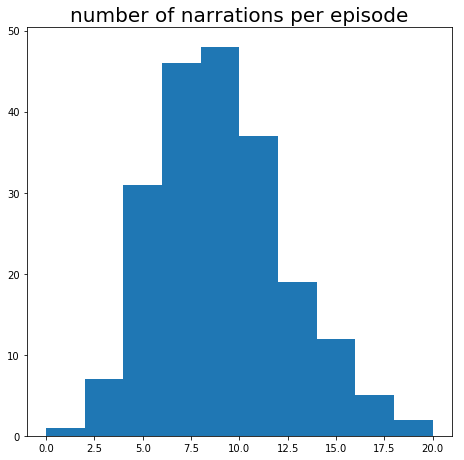}
    \end{subfigure}
    \begin{subfigure}[t]{0.23\textwidth}
    \includegraphics[width=\textwidth]{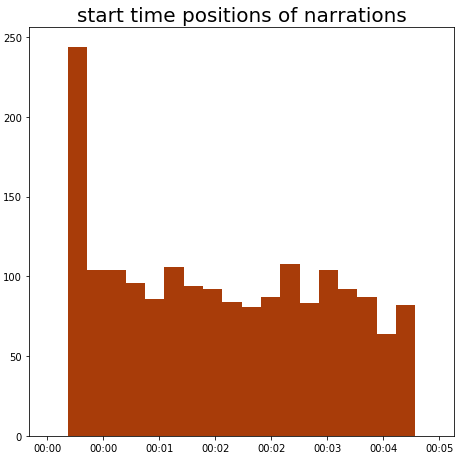}
    \end{subfigure}
    \caption{Histograms of characteristics of Peppa Pig narrations. The length of narrations in sentences and tokens, the start time position within the episode and the number of narrations per episode are listed.}
    \label{fig:narr_length_stats}
\end{figure}

\section{Dataset Preprocessing}\label{appendix:preprocessing}

Each episode was preprocessed as follows:
\begin{itemize}
    \item Double episodes\footnote{most video files contained two episodes} were split to two separate episodes and corresponding subtitles were manually synchronized with the video.
    \item Each episode was trimmed to the correct length. Commonly, episodes start with an ``introduction" part, in which Peppa, the main character, introduces herself and her family. This part, which usually lasts less than two minutes, was manually removed from the episodes of the dataset.
    \item The audio channel was extracted from each episode.
    \item We performed normalization of the text, by manually inspecting it. 
    Since Peppa Pig is a cartoon targeting young children, animal sounds and sound effects are abundant in its dialogues.
    We manually identified and normalized such instances. As an example, all the occurrences of ``oink", ``ooink", ``oinking", ``big oink", ``loud snort" etc. were normalized to ``*pig\_sound*" for text application and to a special noise token (``\{ns\}") for forced alignment purposes (see next step).
    \item The subtitles were aligned with the audio channel in the token level. 
    We used the Penn Forced Aligner (\texttt{p2fa}; \citealt{yuan2008speaker}) for that.
    \item After forced alignment, using \texttt{ffmpeg},\footnote{\texttt{\url{https://www.ffmpeg.org/}}} we extracted frames (snapshots from the video) for every token, at timestamps corresponding to the beginning, the end and the middle of the token timeframe. 
    \item Annotation of the narrator: the sentences uttered by the narrator were semi-manually annotated.
    Some subtitles, especially those that contained the first utterrance of the narrator after a dialogue between the characters, begin with a tag (``[NARRATOR]" -- subtitles are mostly meant for the hearing impaired, since Peppa Pig is aimed at preschoolers). 
    However, the use of this cue is not consistent, and also, it is not used more than once in consecutive narrator subtitles.
    First, we annotated the subtitles beginning with the ``NARRATOR" tag as belonging to the narrator, and a second manual pass was done to ensure that all narrations are properly tagged as such.
\item MFCC features were calculated using the tool \texttt{aubio}.\footnote{\texttt{\url{https://aubio.org/}}}
\item The preprocessing and the annotation was done by the authors, as it is relatively light and does not require expert or crowdsourced annotators.
\end{itemize}

\section{Narration as Video Summary}\label{appendix:narration_as_summary}  

Besides their apparent purpose in books and videos, narrations can also serve as a form of summary.
Consider for example, narrations from two episodes of Peppa Pig along with their corresponding plot summaries (from IMDb)\footnote{\texttt{\url{www.imdb.com}}} and plot sentences (from Wikipedia),\footnote{\texttt{\url{en.wikipedia.org/wiki/List\_of\_Peppa\_Pig\_episodes}}, not available for all episodes.} shown in Figure \ref{fig:narrations_summaries}. 
Clearly, the narrations are more elaborate, but do not fail to convey the main events taking place during the episode. 
Narrations and plot summaries can both be regarded summaries of the video, in different layers of abstraction.

\begin{figure}[th!]
    \centering
    \begin{tabular}{| p{0.45\textwidth} | }
\hline
\small \textbf{narration} \\
\hline
\small It is raining today, so Peppa and George cannot play outside.

Peppa loves jumping in muddy puddles.

George likes to jump in muddy puddles, too.

Peppa likes to look after her little brother, George.

Peppa and George are having a lot of fun.

Peppa has found a little puddle.

George has found a big puddle.

George wants to jump into the big puddle first.

Peppa and George love jumping in muddy puddles.

Peppa and George are wearing their boots.

Mummy and Daddy are wearing their boots.

Peppa loved jumping up and down in muddy puddles.

Everyone loves jumping up and down in muddy puddles.
\\
\hline
\small \textbf{plot summary} \\
\hline
\small
It is raining and Peppa is sad because she cannot go outside. When the rain stops, Peppa and George get to play one of their favourite games - jumping in muddy puddles. Things get very muddy indeed when Mummy and Daddy Pig join in.\\
\hline
\small \textbf{plot} \\
\hline
\small Peppa and George get very muddy after playing their favourite game - Muddy Puddles. \\
\hline
\hline
\small \textbf{narration} \\
\hline
\small Mommy Pig is working on her computer.

Daddy Pig is making soup for lunch.

Mommy Pig has a lot of important work to do.

Peppa and George love to watch Mommy work on the computer.

Oh, dear, the computer is not meant to do that.

Daddy Pig is going to mend the computer.

Daddy Pig has mended the computer.
\\
\hline
\small \textbf{plot summary} \\
\hline
\small
Peppa and George accidentally break Mummy Pig's computer, so Daddy Pig tries to fix it.\\
\hline
\small \textbf{plot} \\
\hline
\small Peppa breaks Mummy Pig's computer while she is working. \\
\hline
\end{tabular}
\caption{Narrations and plot summaries from two episodes, \textit{Episode 1: Muddy Puddles} and \textit{Episode 7: Mummy Pig at Work}.}
\label{fig:narrations_summaries}
\end{figure}

To further explore this idea, we compare the plot summaries and the narrations of the dataset, in order to assess the capabilities of an oracle narrator model as a summarizer.
We work on the full dataset and construct ``summaries" by concatenating all the correct narration sentences for each episode.
Table \ref{table:summarisation} reports ROUGE scores \cite{lin2004rouge} for full length summaries and limited length summaries (75 bytes).\footnote{The evaluation was done using \texttt{pyrouge, available in \url{https://github.com/bheinzerling/pyrouge}}.}
Interestingly, the limited length scores are relatively high, pointing to the fact that the first narration sentence of each episode contains important information for its summary.
However, the low scores in general demonstrate that narration generation is quite a distinct task from summarization, and points to the value of a dataset for this problem.

\begin{table}[h!]
\begin{center}
\begin{tabular}{| l | c | c | c |}
\hline
\multicolumn{4}{|c|}{\textbf{75 bytes}} \\ 
\hline
\textbf{metric} & \textbf{prec} & \textbf{rec} & \textbf{f1}\\
\hline
ROUGE-1 & 19.98 & 19.40 & 19.65\\
ROUGE-2 & 5.61 & 5.48 & 5.53\\
ROUGE-L & 15.32 & 17.87 & 16.47\\
\hline
\multicolumn{4}{|c|}{\textbf{full length}} \\ 
\hline 
\textbf{metric} & \textbf{prec} & \textbf{rec} & \textbf{f1}\\
\hline
ROUGE-1 & 31.74 & 11.88 & 16.42\\
ROUGE-2 & 6.27 & 2.37 & 3.23\\
ROUGE-L & 28.22 & 10.55 & 14.58\\
\hline
\end{tabular}
\end{center}
\caption{ROUGE scores (precision, recall and F1) comparing the plot summaries to the corresponding narration sentences of each episode.}
\label{table:summarisation}

\end{table}

\end{document}